# Mind the Noise: Sensitivity of Transformer-based Interaction-Aware Trajectory Prediction Models to Noisy Data

Shahab Salehi, Luca Lusvarghi, Miguel Sepulcre, Javier Gozalvez
Networked Systems Lab, Universidad Miguel Hernandez de Elche, Elche, Spain
Email: {ssalehi, llusvarghi, msepulcre, j.gozalvez}@umh.es

***Abstract*—Trajectory prediction allows autonomous vehicles to anticipate the future behavior of surrounding objects (or agents) and, accordingly, maximize the safety and efficiency of their driving. State-of-the-art Transformed-based interaction-aware trajectory prediction models, which rely on attention mechanisms to capture multi-agent interactions and maximize prediction accuracy, are commonly trained and evaluated on long-range high-quality datasets. These datasets are typically obtained by aggregating data from multiple vehicles or drones and removing any object detection or tracking noise offline. Yet, information about a surrounding object's state (its position, speed, heading) is far from being noiseless in real-world deployments. Object state estimation is affected by perception uncertainties and localization errors that can be particularly large for objects received via Vehicle-to-Everything (V2X) communications. In this paper, we analyze the impact of noisy object state information on the trajectory prediction accuracy of a state-of-the-art Transformer-based interaction-aware trajectory prediction model. Our study demonstrates that trajectory prediction accuracy can rapidly deteriorate as the noise intensity increases. Numerical results show that the prediction accuracy can reduce by a 1.3x factor under small noise levels and by as much as a 3.9x factor under the highest (yet realistic) noise conditions. These findings reveal the strong sensitivity of trajectory prediction models to noisy data, underscoring the need for more realistic training and evaluation datasets as well as noise mitigation strategies.**

## I. Introduction

Autonomous vehicles rely on trajectory prediction models to continuously forecast the evolution of the surrounding environment, in particular, the behavior of nearby agents or objects[1] (vehicles, pedestrians, etc.). Anticipating their future behavior is fundamental for downstream planning and control modules, as it enables autonomous vehicles to detect and avoid potential collisions, comply with traffic rules, and ultimately improve driving safety and efficiency.

Trajectory prediction consists of forecasting the evolution of a target object's state (i.e., its position, velocity, and heading) over a desired prediction horizon, given a sequence of historical observations describing its past behavior. It can either rely on model-driven approaches based on kinematic models coupled with state estimation techniques (e.g., Kalman filters), or data-driven approaches that leverage AI to learn the objects' mobility patterns directly from data. State-of-the-art trajectory prediction models such as DeMo++ [1], SEPT [2], and FutureNet-LOF [3] are based on powerful AI architectures – primarily Transformers – inherited from world models [4]. They integrate the target object's history with rich contextual information, including road topology, traffic rules, and the mobility of other surrounding agents, to capture the driving environment's spatio-temporal evolution and improve trajectory prediction accuracy. Access to contextual cues describing the mobility of surrounding agents is particularly important, as highlighted in [5], since a target object's future trajectory can be strongly influenced by multi-agent interactions (also referred to as agent-to-agent or object-to-object interactions). Accordingly, recent studies have shown that trajectory prediction accuracy improves as the ego-vehicle has access to information about a larger number of surrounding agents, especially in high-density scenarios [6].

State-of-the-art Transformer-based interaction-aware trajectory prediction models are hence commonly trained and evaluated on datasets collected using fleets of vehicles or drones, such as the Argoverse 2 Motion Forecasting dataset [7] and the INTERACTION dataset [8], respectively. These datasets rely on offline post-processing pipelines to: (i) aggregate observations from multiple viewpoints and collect a larger number of surrounding objects, including objects beyond the line-of-sight range of the ego-vehicle's onboard sensors; and (ii) remove detection and tracking noise to obtain high-quality object state information. However, in real-world deployments, information about surrounding objects collected in real time can be noisy. For example, the position, speed, and heading of objects detected with onboard sensors can be affected by a distance-dependent perception noise due to environmental effects (e.g., weather and illumination conditions) as well as onboard sensors hardware and software limitations. Information about objects that the ego-vehicle cannot locally detect with its onboard sensors and that, therefore, it receives from other vehicles or roadside infrastructure nodes via Vehicle-to-Everything (V2X) communications can also be particularly noisy. In this case, the positions of objects are typically encoded with respect to the position of the transmitter, and localization errors at the transmitter can result in a large positioning noise at the receiver. In this context, it is critical to investigate whether state-of-the-art Transformer-based interaction-aware trajectory prediction models are resilient to noisy object state information, and whether they can deliver the trajectory prediction accuracy levels necessary for safe connected and automated driving in real-world deployments.

This paper advances the state-of-the-art by analyzing for the first time the impact that noisy object state information – retrieved by onboard sensors or received via V2X communications – has on the prediction accuracy of

This work was supported by the European Union under the 2023 MSCA Postdoctoral Fellowship program (project no. 101153845), MICIU/AEI/10.13039/501100011033 (PID2023-150308OB-I00), and by the Generalitat Valenciana (CIAICO/2024/167).

[1]The terms agents and objects are used interchangeably throughout this work.

Transformer-based interaction-aware trajectory prediction models. To this aim, we consider the state-of-the-art DeMo++ trajectory prediction model and the Argoverse 2 Motion Forecasting dataset and study the impact of noise levels that reflect realistic onboard perception and V2X-related uncertainties. Our study demonstrates that trajectory prediction accuracy degrades significantly as noise intensity increases. Numerical results show that accuracy can deteriorate by a 1.3x factor even under negligible noise levels, with degradation reaching up to a 3.9x factor when the highest (yet realistic) noise levels are considered. These findings highlight the strong sensitivity of state-of-the-art trajectory prediction models to noisy input data, and underscore the need for more realistic training and evaluation datasets that capture typical real-world object state estimation uncertainties, as well as for effective noise mitigation techniques.

## II. Trajectory Prediction

### A. *Trajectory Prediction Models*

Early trajectory prediction models were primarily based on model-driven techniques such as Kalman filtering. While computationally efficient, Kalman filters rely on simplified modelling assumptions (e.g., constant velocity or acceleration), do not take into account interactions between agents, and generate a single predicted trajectory per target object. As a result, they struggle in real-world driving scenarios where objects' mobility is highly nonlinear, interaction-dependent, and inherently multimodal since each object can exhibit multiple plausible future behaviors.

To overcome these limitations, data-driven (or learning-based) trajectory prediction approaches that rely on AI-based architectures to learn the objects' mobility directly from data were introduced. First data-driven trajectory prediction models employed Convolutional Neural Networks (CNNs) and Recurrent Neural Networks (RNNs). CNN-based trajectory prediction models are particularly effective at capturing local spatio-temporal patterns and spatial correlations in mobility data, such as short-term interactions between nearby agents. However, CNNs have a limited ability to model long-range temporal dependencies that restricts their effectiveness in real-world deployments. RNNs can explicitly model long-range temporal dependencies in the objects' mobility. This is the case of Long Short-Term Memory (LSTM) networks that use gating mechanisms and have been widely employed in trajectory prediction. Despite their advantages, RNNs and LSTMs remain difficult to train on very long input sequences and are affected by vanishing and exploding gradient problems that can limit their scalability and performance in complex driving scenarios [9].

More recently, attention mechanisms have emerged as a key component of state-of-the-art trajectory prediction models. Unlike RNN-based approaches, attention mechanisms can directly and selectively access all elements of an input sequence (i.e., past object behavior) and its surrounding context, facilitating the learning of long-range dependencies and complex interaction patterns. Accordingly, trajectory prediction models based on Transformer architectures and attention mechanisms exhibit higher prediction accuracies than their RNN-based counterparts [9]. By explicitly modeling agent-to-agent and agent-to-lane relationships across extended temporal horizons, Transformer-based interaction-aware models such as SEPT [2], FutureNet-LOF [3], SIMPL [10], and DeMo++ [1] achieve state-of-the-art performance on trajectory prediction benchmarks.

### B. *Datasets*

AI-based trajectory prediction models, including those based on Transformer architectures and attention mechanisms, are trained on trajectory prediction datasets. These datasets typically consist of a collection of scenes, where each scene is a $T$ seconds long recording of the ego-vehicle and its surrounding objects' trajectory. The trajectory of each object within the scene is a temporal sequence of its state (i.e., position, velocity, and heading), periodically sampled every $\tau$ seconds. The first $T_H$ seconds represent the object's past behavior (or history) and are the input to the trajectory prediction model. The last $T_F$ seconds represent the object's ground-truth future trajectory and are the supervision signal used to train the models.

Existing trajectory prediction datasets can be broadly categorized into synthetic and real-world datasets. Synthetic datasets are generated using traffic simulators like SUMO or CARLA and are valuable for rapid prototyping and testing under controlled and reproducible driving scenarios. However, they are typically generated with simplified car-following models that may not fully capture the complexity of real-world driving dynamics. On the other hand, real-world datasets are more challenging and expensive to collect, but they provide valuable data that reflects the complex, nonlinear, and interaction-rich dynamics of real-world driving environments. For this reason, real-world trajectory prediction datasets are the preferred choice for the training and evaluation of AI-based trajectory prediction models.

Early real-world datasets, such as nuScenes [11], were collected using the onboard sensors of a single vehicle. Due to the line-of-sight and range limitations of onboard sensors, these datasets included information on a limited number of objects within the driving scenario. More recent datasets are collected by aggregating data recorded by fleets of vehicles, drones, or roadside infrastructure sensors. The aggregation of multi-view observations significantly increases the density of objects per scene, providing richer contextual information—a critical requirement for maximizing trajectory prediction accuracy [12]. State-of-the-art real-world datasets collected by aggregating multi-view data include Argoverse 1 and Argoverse 2 Motion Forecasting [7] datasets, the Waymo Open Motion Dataset [13], Lyft [14], and INTERACTION [8]. All these datasets provide highly curated object trajectories by removing real-world perception and tracking uncertainties offline. The Argoverse 2 Motion Forecasting dataset excels with its vast scale of 250,000 scenarios and a 250 m spatial horizon which significantly exceeds the range of nuScenes, and Waymo. This extended coverage across six cities and 2,220 km of unique roads provides the necessary diversity and depth for training and evaluating state-of-the-art trajectory prediction models.

## III. Experiments Setup

We investigate the resilience of state-of-the-art trajectory prediction models to noisy object state information using DeMo++, a Transformer-based interaction-aware trajectory prediction model, and the Argoverse 2 Motion Forecasting dataset (or Argoverse 2).

### A. *DeMo++*

DeMo++ [1] is an open-source model ranked in the top ten of the Argoverse 2 Motion Forecasting leaderboard (at the time of writing). Its architecture is built upon the Transformer framework, utilizing attention mechanisms to effectively capture temporal dependencies and complex social interactions between agents. DeMo++ is a multi-modal trajectory prediction model that generates six distinct predicted trajectories for each target object. Each predicted trajectory is assigned a confidence score. DeMo++ employs separate attention modules to capture agent-to-agent interactions, which describe the influence of agents on each other's behavior, and lane-to-agent interactions, which encode the constraints imposed by road geometry and lane topology. This explicit separation enables the model to jointly reason about agent interactions and map semantics, both of which are critical for accurate trajectory prediction in complex driving scenarios. A key characteristic of DeMo++ is its motion decoupling strategy, whereby future trajectories are represented by distinct directional intentions and dynamic states. These distinct representations are then integrated to form a complete long-term prediction. This step-wise prediction process allows the model to capture global motion intent over longer time horizons while maintaining sensitivity to fine-grained temporal dynamics at each prediction step, resulting in improved prediction stability and accuracy. Furthermore, DeMo++ incorporates cross-scene intention interactions that allow the model to associate motion intentions across different driving scenes during training. By encouraging cross-scene consistency in predicted motion patterns, this mechanism aims to improve the robustness and generalization capability of the model when encountering diverse road layouts. This decoupled design allows DeMo++ to generalize across complex, real-world driving environments with varying traffic conditions, achieving high levels of precision across the Argoverse 2 benchmark.

## B. *Argoverse 2*

Argoverse 2 [7] is a large-scale, widely adopted dataset for the training, evaluation, and benchmarking of trajectory prediction models. The dataset consists of approximately 250,000 driving scenes recorded across 6 major U.S. cities – Austin, Detroit, Miami, Pittsburgh, Palo Alto, and Washington, D.C. – employing a fleet of vehicles equipped with two rooftop lidar sensors, seven ring cameras, and two front-facing stereo cameras. Each driving scene is $T$ = 11 seconds long and includes annotated trajectories of the ego-vehicle and all objects observed in the scene. The ego-vehicle is one of the vehicles from the recording fleet. Other vehicles from the recording fleet are encoded as objects. Each object trajectory is divided into a past observation window that covers its first $T_H$ = 5 seconds and a future prediction horizon that spans its last $T_F$ = 6 seconds. The dataset includes 10 different object types, such as vehicles, pedestrians, and cyclists, and the trajectory of each object consists of a sequence of data points representing the object's state (i.e., its position, speed, and heading) time evolution, sampled at 10 Hz (every $\tau$ = 100 ms). At each timestep, each object's state information is estimated through a combination of GPS-based localization, onboard sensing, and offline sensor fusion and post-processing techniques that ultimately allow the recording of high-quality, smooth object trajectories. A distinctive feature of Argoverse 2 is its extended spatial coverage and rich contextual information. By aggregating – offline – observations collected from a fleet of vehicles, the dataset includes objects located well beyond the ego-vehicle sensors' line-of-sight range. In particular, Argoverse 2 includes objects located up to 250 m from the ego vehicle. This long-range information horizon is key to allow trajectory prediction models to deal with complex multi-agent interactions, making the dataset particularly suitable for the design and evaluation of state-of-the-art trajectory prediction models. We should though note that, in real-world deployments, a vehicle can collect information about objects beyond its onboard sensors' perception range only through V2X communications.

## C. *Noise Model*

Information about surrounding objects collected in real-time may be noisy, and it is critical to understand the impact that noisy object state information – collected by onboard sensors or received via V2X communications – may have on the trajectory prediction accuracy of Transformer-based interaction-aware trajectory prediction models such as DeMo++. We perturb the state of the dataset's objects with two different types of noise that reflect realistic object state estimation uncertainties. At each timestep, we add a zero-mean Gaussian noise $N_S \sim (0, \sigma_S)$ to the position, speed, and heading of objects detected by the ego-vehicle's onboard sensors. We will refer to this type of noise as local perception noise. This local perception noise captures typical onboard perception uncertainties due to environmental effects (bad weather and illumination conditions) and onboard sensors hardware and software limitations. In line with [15], we define the standard deviation $\sigma_S$ of local perception noise as $\sigma_S = 0.001 \times D$,where $D$ indicates the distance between the object and the ego-vehicle. Position noise is applied directly in meters (m), while speed and heading noise are derived from the same distance-dependent uncertainty and expressed in their respective units (m/s and radians). At each timestep, we model the probability that an object $o_i$ can be locally detected by the ego-vehicle $V_e$ with its onboard sensors through a detection probability $P(d_i)$. If there is another object $o_j$ blocking the direct line-of-sight between $o_i$ and $V_e$, then $o_i$ cannot be detected by $V_e$'s onboard sensors and $P(d_i) = 0$. If there is no object blocking the direct line-of-sight between $o_i$ and $V_e$, then $o_i$ can be detected by $V_e$'s onboard sensors with a probability $P(d_i)$ that depends on the distance $d_i$ between $o_i$ and $V_e$:

$$P(d_i) = \begin{cases} 1 & d_i \leq 110 \\ \dfrac{200 - d_i}{200 - 110} & 110 < d_i < 200 \\ 0 & d_i \geq 200 \end{cases} \quad (1)$$

The detection probability in eq. (1) is in line with the detection probability that characterizes a vehicle's onboard sensors perception range in other studies [16].

We perturb the position of objects that the ego-vehicle cannot detect through its onboard sensors and that it receives through V2X communications with a zero-mean Gaussian noise $N_{V2X} \sim (0, \sigma_{V2X})$. We will refer to this type of noise on the object's position as V2X noise in the rest of this paper. The standard deviation of the V2X noise $\sigma_{V2X}$ captures the intensity of GNSS-related localization errors at the transmitter that can ultimately affect the estimation of the received object's position at the ego-vehicle.

# IV. Numerical Results

## A. *Qualitative Analysis*

Fig. 1(a)-(f) qualitatively illustrate the impact of noisy object state information on the accuracy of trajectory predictions reported by DeMo++ for a target object extracted from a representative Argoverse 2 scene. The impact is

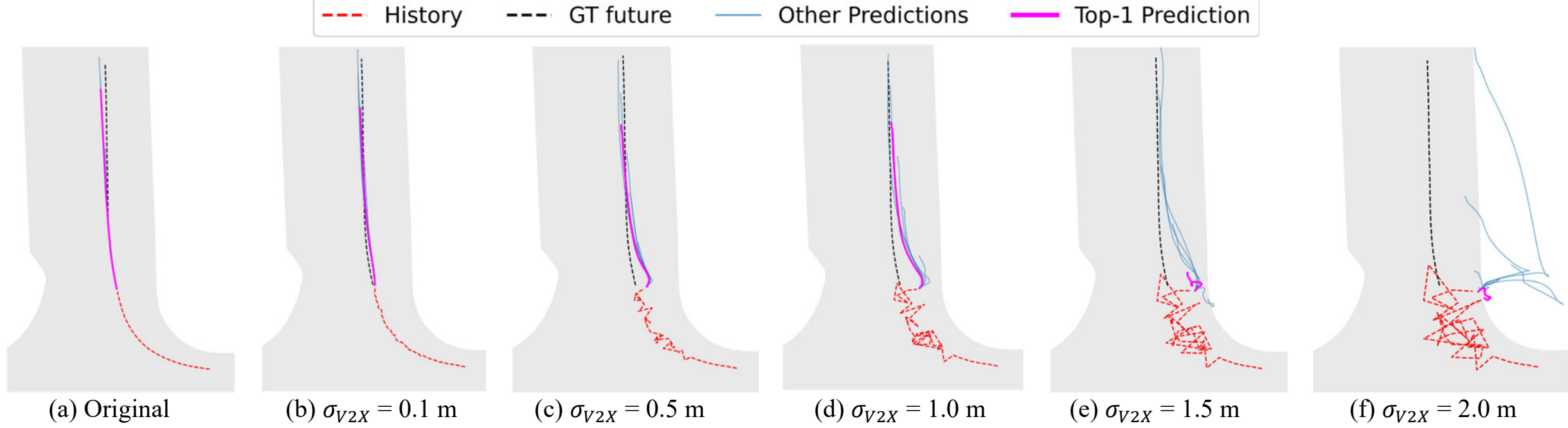

Fig. 1. DeMo++ output under varying V2X noise levels ($\sigma_{V2X}$).

represented under increasing levels of V2X noise ($\sigma_{V2X}$). Each figure shows the target object's history (i.e., its past trajectory over the last $T_H$ seconds), its ground-truth (GT) future trajectory (over the next $T_F$ seconds), and the 6 trajectories predicted by DeMo++. The predicted trajectory associated with the highest confidence score is referred to as the top-1 predicted trajectory.

Fig. 1(a) shows DeMo++ predicted trajectories in the original Argoverse 2 scene, with no noise ($\sigma_S = \sigma_{V2X} = 0.0$ m). In this case, all 6 DeMo++ predicted trajectories closely match the target object's ground-truth future trajectory. From Fig. 1(b) onwards, the target object's past trajectory is perturbed with two different types of noise. Specifically, the target object's state estimates are perturbed with local perception noise ($N_S$, Section III.C) during the first 1.5 seconds of its past trajectory, when the target object lies within the ego-vehicle's perception range. Given a maximum local detection range of 200 m (eq. (1) in Section III.C), the maximum local perception noise standard deviation is $\sigma_S =$ 0.2 m. During the last 3.5 seconds, when the target object is beyond the ego-vehicle's perception range, its estimated position is affected by V2X noise ($N_{V2X}$, Section III.C). The range of V2X noise levels analyzed in Fig. 1 (from $\sigma_{V2X}$ = 0.1 m to $\sigma_{V2X}$ = 2.0 m) is in line with real-world GNSS-related localization errors [17].

Fig. 1 shows that the trajectory prediction accuracy rapidly deteriorates as the V2X noise level increases. In Fig. 1(b) and Fig. 1(c), under low V2X noise levels, DeMo++ predictions do not significantly deviate from the ground-truth reference and are comparable with the original noiseless scenario of Fig. 1(a). DeMo++ prediction errors begin to increase from intermediate V2X noise levels (Fig. 1(d)) with sharp degradations observed under high V2X noise levels (Fig. 1(e) and Fig. 1(f)). In the latter case, predicted trajectories significantly deviate from the ground-truth reference, even extending beyond the road boundaries in Fig. 1(f). Notably, the top-1 predicted trajectory is often the least accurate in high V2X noise scenarios.

## B. Quantitative Analysis

Fig. 1 provides a clear and visual example of the sensitivity of a state-of-the-art Transformer-based interaction-aware trajectory prediction model such as DeMo++ to noisy object state information. We now quantitatively analyze such sensitivity by evaluating DeMo++ on the entire Argoverse 2 dataset using three standard metrics commonly adopted in trajectory prediction benchmarks: minimum Final Displacement Error ($\text{minFDE}_K$), minimum Average Displacement Error ($\text{minADE}_K$), and Miss Rate ($\text{MR}_K$). The $\text{minFDE}_K$ metric is defined as the Euclidean distance between the final point of the best predicted trajectory and the final point of the ground-truth future trajectory. A multi-modal trajectory prediction model, such as DeMo++, outputs $K$ predicted trajectories per object. The best predicted trajectory is the one with the minimum final point distance among the $K$ predictions. The best trajectory is used to calculate all other metrics. The $\text{minADE}_K$ metric is defined as the average Euclidean distance between all points of the best predicted trajectory and the ground-truth trajectory. Finally, the $\text{MR}_K$ metric corresponds to the fraction of target objects with a $\text{minFDE}_K$ larger than 2.0 m.

Table I reports the $\text{minFDE}_K$, $\text{minADE}_K$, and $\text{MR}_K$ metrics obtained with $K$ = 1 (Table I(a)) and $K$ = 6 (Table I(b)) considering the same V2X noise levels employed by the qualitative analysis. When $K$ = 1, metrics are computed considering only the top-1 predicted trajectory, i.e., the predicted trajectory associated with the highest confidence score. When $K$ = 6, metrics are computed considering all six predicted trajectories generated by DeMo++. The comparison between Table I(a) and Table I(b) shows that the predicted trajectory with the highest confidence score is generally not the most accurate one. Metrics computed considering only the top-1 predicted trajectory ($K$ = 1) are worse (i.e., larger) than their counterparts computed over all six predicted trajectories ($K$ = 6), also when the original Argoverse 2 data is considered. This is a well-known trend in the literature and highlights the difficulty these models face in accurately ranking their predicted trajectories.

Both Table I(a) and Table I(b) numerically demonstrate the strong sensitivity of DeMo++ to noisy object state information. Increasing V2X noise levels lead to a monotonic trajectory prediction accuracy degradation across all evaluation metrics. In particular, Table I(a) shows that the

TABLE I. DEMO++ TRAJECTORY PREDICTION ACCURACY UNDER THE PRESENCE OF NOISE.

| $\sigma_{V2X}$ [m] | $\text{minFDE}_1$ [m] | $\text{minADE}_1$ [m] | $\text{MR}_1$ |
|---|---|---|---|
| Original – 0.0 | 3.700 | 1.500 | 0.550 |
| 0.1 | 4.480 | 1.930 | 0.618 |
| 0.2 | 4.700 | 2.071 | 0.637 |
| 0.5 | 5.444 | 2.595 | 0.666 |
| 1.0 | 6.732 | 3.427 | 0.691 |
| 1.5 | 7.896 | 4.112 | 0.703 |
| 2.0 | 8.782 | 4.628 | 0.710 |

(a) $K$ = 1

| $\sigma_{V2X}$ [m] | $\text{minFDE}_6$ [m] | $\text{minADE}_6$ [m] | $\text{MR}_6$ |
|---|---|---|---|
| Original – 0.0 | 1.120 | 0.610 | 0.120 |
| 0.1 | 1.483 | 0.947 | 0.176 |
| 0.2 | 1.565 | 1.025 | 0.187 |
| 0.5 | 1.856 | 1.303 | 0.229 |
| 1.0 | 2.429 | 1.760 | 0.277 |
| 1.5 | 2.913 | 2.106 | 0.305 |
| 2.0 | 3.290 | 2.373 | 0.321 |

(b) $K$ = 6

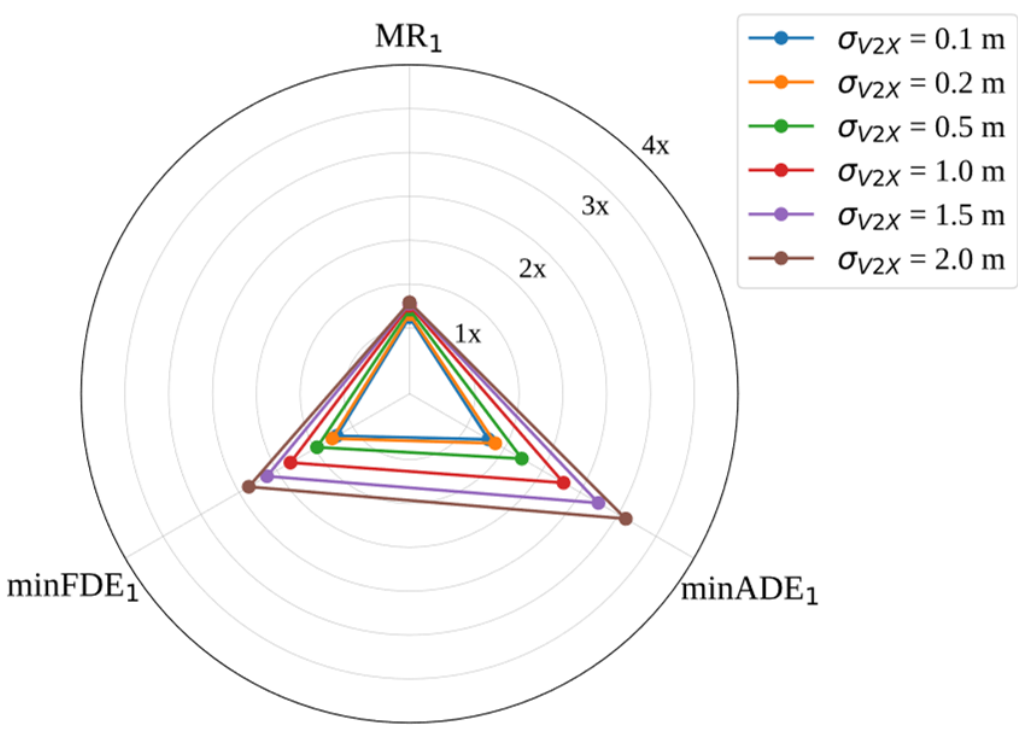

Fig. 2. Relative increase of the $\text{minADE}_1$, $\text{minFDE}_1$, and $\text{MR}_1$ metrics with respect to the original noiseless scenario under increasing V2X noise levels.

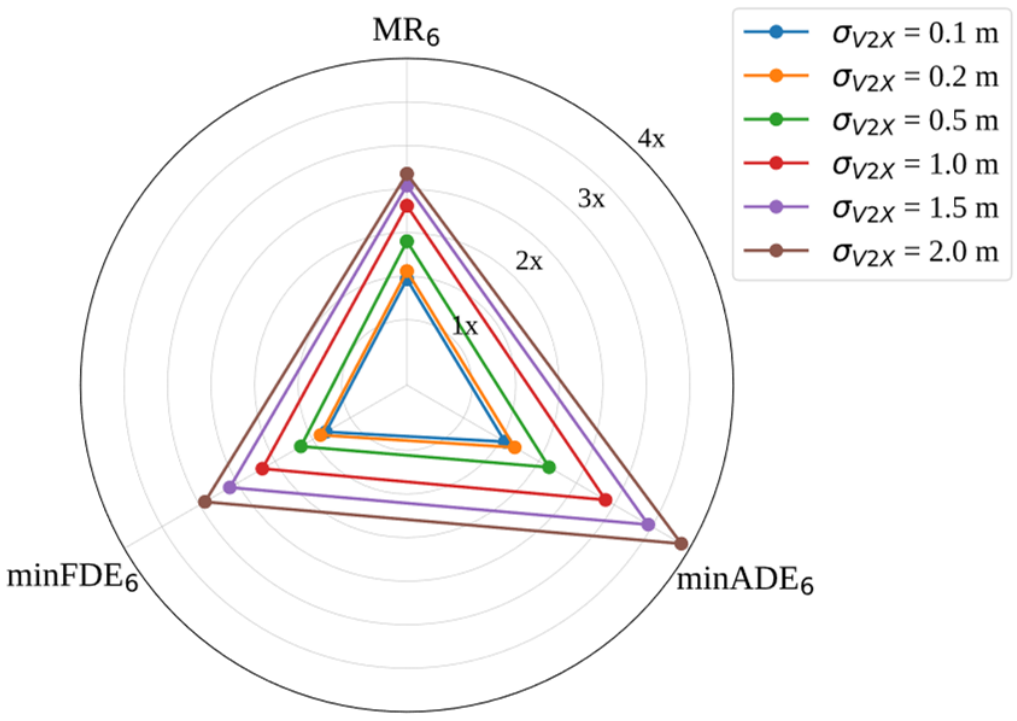

Fig. 3. Relative increase of the $\text{minADE}_6$, $\text{minFDE}_6$, and $\text{MR}_6$ metrics with respect to the original noiseless scenario under increasing V2X noise levels.

$\text{minFDE}_1$ increases from the 3.700 m reference obtained on the original Argoverse 2 data up to 8.782 m when $\sigma_{V2X}$ = 2.0 m – an almost threefold increase. A similar increase, although with smaller absolute values, is observed in the $\text{minFDE}_6$ case reported in Table I(b). The $\text{minFDE}_6$ increases from 1.120 m (in the noiseless setting) to 3.290 m.

The relative increase, with respect to the original noiseless setting, of the three metrics under growing V2X noise levels is reported in Fig. 2 and Fig. 3 for $K$ = 1 and $K$ = 6, respectively. Fig. 2 shows that the $\text{minFDE}_1$ and $\text{minADE}_1$ metrics can increase by approximately 1.3x, even when the V2X noise intensity is low ($\sigma_{V2X}$ = 0.1 m). Under the highest noise intensity ($\sigma_{V2X}$ = 2.0 m), the $\text{minFDE}_1$ and $\text{minADE}_1$ exhibit a 2.4x and 3.1x increase, respectively. In contrast, the $\text{MR}_1$ shows a relative increase always below 1.3x for all considered V2X noise levels. This is the case since the $\text{MR}_1$ metric attains pretty large values (> 0.5) also in the original noiseless scenario, as shown by Table I. The largest trajectory prediction accuracy degradation is observed with $K$ = 6 in Fig. 3. The relative increase of all metrics ranges from 1.3x to 3.9x (observed in the $\text{minADE}_6$ case). It is worth highlighting that, in both Fig. 2 and Fig. 3, $\text{minADE}_K$ exhibits a larger relative increase than $\text{minFDE}_K$, across all V2X noise levels. This indicates that noisy object state information affects the overall shape and temporal consistency of predicted trajectories more severely than their final endpoints.

## V. Conclusions

Trajectory prediction is fundamental for connected and automated driving, and Transformer-based interaction-aware trajectory prediction models provide, to date, the best prediction accuracy. However, these attention-based models are usually trained and evaluated on long-range high-quality datasets processed to remove detection and tracking noise, while information collected in real-world deployments is inherently noisy. This study demonstrates the significant impact that noisy object state information – collected by onboard sensors or received using V2X communications – can have on the trajectory prediction accuracy. Our analysis shows that trajectory prediction accuracy can degrade by a factor of 1.3x even under small noise levels, and by up to 3.9x under more severe noise conditions. These findings indicate that the high accuracy reported by current state-of-the-art models may not directly translate to real-world settings where object state estimation is inherently noisy. They also underscore potential limitations of existing training and evaluation practices based on highly curated datasets, calling for the development of more realistic datasets that account for realistic object state estimation uncertainties, as well as for the design of robust noise mitigation techniques. Addressing these challenges is essential to bridge the gap between offline benchmarking and real-world deployment performance.

## References

[1] B. Zhang et al., “DeMo++: Motion Decoupling for Autonomous Driving,” arXiv:2507.17342, July 2025.

[2] Lan, Z et al., “SEPT: Towards Efficient Scene Representation Learning for Motion Prediction,” in *Proc. ICLR 2024*, Vienna, Austria, 2024.

[3] M. Wang et al., “FutureNet-LoF: Joint Trajectory Prediction and Lane Occupancy Field Prediction with Future Context Encoding,” in *Proc. ICRA 2024*, Yokohama, Japan, 2024, pp. 8841-8848.

[4] J. Ding et al., “Understanding World or Predicting Future? A Comprehensive Survey of World Models,” *ACM Computing Surveys*, vol. 58, no. 3, 2026, pp. 1-38.

[5] N. A. Madjid et al., “Trajectory prediction for autonomous driving: Progress, limitations, and future directions,” *Elsevier Information Fusion*, vol. 126, Feb. 2026, pp. 1-59.

[6] F. Diehl et al., “Graph neural networks for modelling traffic participant interaction,” in *Proc. IEEE IV 2019*, Paris, France, 2019, pp. 695–701.

[7] B. Wilson et al., “Argoverse 2: Next Generation Datasets for Self-Driving Perception and Forecasting,” arXiv:2301.00493, Jan. 2023.

[8] W. Zhan et al., “INTERACTION Dataset: An INTERnational, Adversarial and Cooperative moTION Dataset in Interactive Driving Scenarios with Semantic Maps,” arXiv:1910.03088, Sep. 2019.

[9] Y. Huang, et al., “A Survey on Trajectory-Prediction Methods for Autonomous Driving,” *IEEE Transactions on Intelligent Vehicles*, vol. 7, no. 3, Sep. 2022, pp. 652–674.

[10] L. Zhang, P. Li, S. Liu, and S. Shen, “SIMPL: A Simple and Efficient Multi-Agent Motion Prediction Baseline for Autonomous Driving,” *IEEE Robot. Autom. Lett.*, vol. 9, no. 4, Apr. 2024, pp. 3767–3774.

[11] H. Caesar et al., “nuscenes: A multimodal dataset for autonomous driving,” in *Proc. IEEE/CVF CVPR 2020*, Seattle, WA, USA, June 2020, pp. 11621–11631.

[12] H. Zhang et al., “OOSTraj: Out-of-Sight Trajectory Prediction With Vision-Positioning Denoising,” in *Proc. IEEE/CVF CVPR 2024*, Seattle, WA, USA, Jun. 2024, pp. 14802–14811.

[13] S. Ettinger et al., “Large Scale Interactive Motion Forecasting for Autonomous Driving: The Waymo Open Motion Dataset,” in *Proc. IEEE/CVF ICCV 2021*, Montreal, Canada, Oct. 2021, pp. 9710–9719.

[14] J. Houston et al., “One Thousand and One Hours: Self-driving Motion Prediction Dataset,” in *Proc. CoRL 2020*, Cambridge, MA, USA, Jan. 2021, pp. 409–418

[15] A. Mohammadisarab et al., “How the Fusion of Onboard Sensors and V2X Data can Improve (or not) the Cooperative Perception of Connected Automated Vehicles,” in *Proc. VTC2025-Spring*, Oslo, Norway, 2025, pp. 1-5.

[16] G. Thandavarayan, et al., “Generation of Cooperative Perception Messages for Connected and Automated Vehicles,” *IEEE Transactions on Vehicular Technology*, vol. 69, no. 12, Dec. 2020, pp. 16336-16341.

[17] M. M. S. Alghananim and W. Y. Ochieng, “Maximum non-bounded difference method for overbounding Global Navigation Satellite System errors,” *GPS Solutions*, vol. 29, no. 1, 2025, pp. 1–14.